\documentclass[conference]{IEEEtran}
\IEEEoverridecommandlockouts
\usepackage{pgfplots}
\usepgfplotslibrary{groupplots} % <<-- C'EST CETTE LIGNE QUI MANQUE
\pgfplotsset{compat=1.18}
\usepackage{tikz}
\usepackage{pgfplots}
\pgfplotsset{compat=1.18} % Utilisation du moteur moderne de pgfplots
\usepackage{amsmath} % Pour les notations mathématiques (optionnel, mais bonne pratique)
\usepackage{xcolor} % Pour les couleurs personnalisées
\usepackage{cite}
\usepackage{amsmath,amssymb,amsfonts}
\usepackage{algorithmic}
\usepackage{graphicx}
\usepackage{textcomp}
\usepackage{xcolor}
\usepackage{tabularx}
\usepackage{booktabs} % Recommandé pour les tableaux propres sans lignes verticales
\usepackage{tabularx} % Nécessaire
\usepackage{multirow} % Nécessaire
\usepackage{caption} % Optionnel, pour une meilleure gestion des légendes
\usepackage{pgfplotstable}
\usepackage{pgfplots}
\pgfplotsset{compat=1.18}
\usepackage{graphicx}
\usepackage{subcaption}

\usepackage[
  colorlinks=true,
  linkcolor=blue,
  citecolor=blue,
  urlcolor=blue
]{hyperref}

\def\BibTeX{{\rm B\kern-.05em{\sc i\kern-.025em b}\kern-.08em
    T\kern-.1667em\lower.7ex\hbox{E}\kern-.125emX}}
    
\begin{document}

\title{Spectral Aliasing Pretext: A novel task for Self-Supervised fault diagnosis in rotating machinery}   %\\
\author{
    \IEEEauthorblockN{Victor Gialis\IEEEauthorrefmark{1}\IEEEauthorrefmark{2}, Maxime Metz\IEEEauthorrefmark{2}\IEEEauthorrefmark{3}\IEEEauthorrefmark{4}, David Esteve \IEEEauthorrefmark{2}\IEEEauthorrefmark{3}, Abdenour Soualhi\IEEEauthorrefmark{1}}
    \IEEEauthorblockA{\IEEEauthorrefmark{1}LASPI, Univ. Jean Monnet, Roanne, 42300, France}
    \IEEEauthorblockA{\IEEEauthorrefmark{2}Pellenc ST, Applied Research Group, Pertuis, 84120, France}
    \IEEEauthorblockA{\IEEEauthorrefmark{3}LabCom Aioly, Artificial Intelligence and Optics Laboratory, Montpellier, 34196, France}
    \IEEEauthorblockA{\IEEEauthorrefmark{4}IMBE, Univ. Aix-Marseille, UMR CNRS IRD Univ. Avignon, Site de l’Etoile Marseille, France
    \\\{victor.gialis@univ-st-etienne.fr, m.metz@pellencst.com, d.esteve@pellencst.com, abdenour.soualhi@univ-st-etienne.fr\}}
}

\maketitle

\begin{abstract}
Deep learning is a new way for machinery fault diagnosis but requires extensive labeled data, a scarce resource in industrial settings. We propose Spectral Aliasing Pretext (SAP), a self-supervised learning method that pretrains models on unlabeled vibration data by exploiting spectral aliasing. We deliberately undersample signals to create folded spectrum, then train a Transformer to reconstruct the original unfolded spectrum. This pretext task forces the model to learn frequency-domain invariants characteristic of mechanical faults, without potentially destructive augmentations. Experiments on the CWRU dataset show that SAP learns stable and highly discriminative representations. In a linear probing setting, SAP quickly achieves very high classification performance with only a small fraction of labeled data and low variance. In contrast, full fine-tuning, including fully supervised training, does not lead to more stable or better results. Overall, these findings suggest that SAP combined with linear probing can be more effective and reliable than fully supervised training for fault diagnosis with limited labeled data.

\end{abstract}

\begin{IEEEkeywords}
frequency domain, transformer, self-supervised learning, fault diagnosis, vibration data
\end{IEEEkeywords}

\section{Introduction}
\begin{figure*}
    \centering
    \includegraphics[width=1\linewidth]{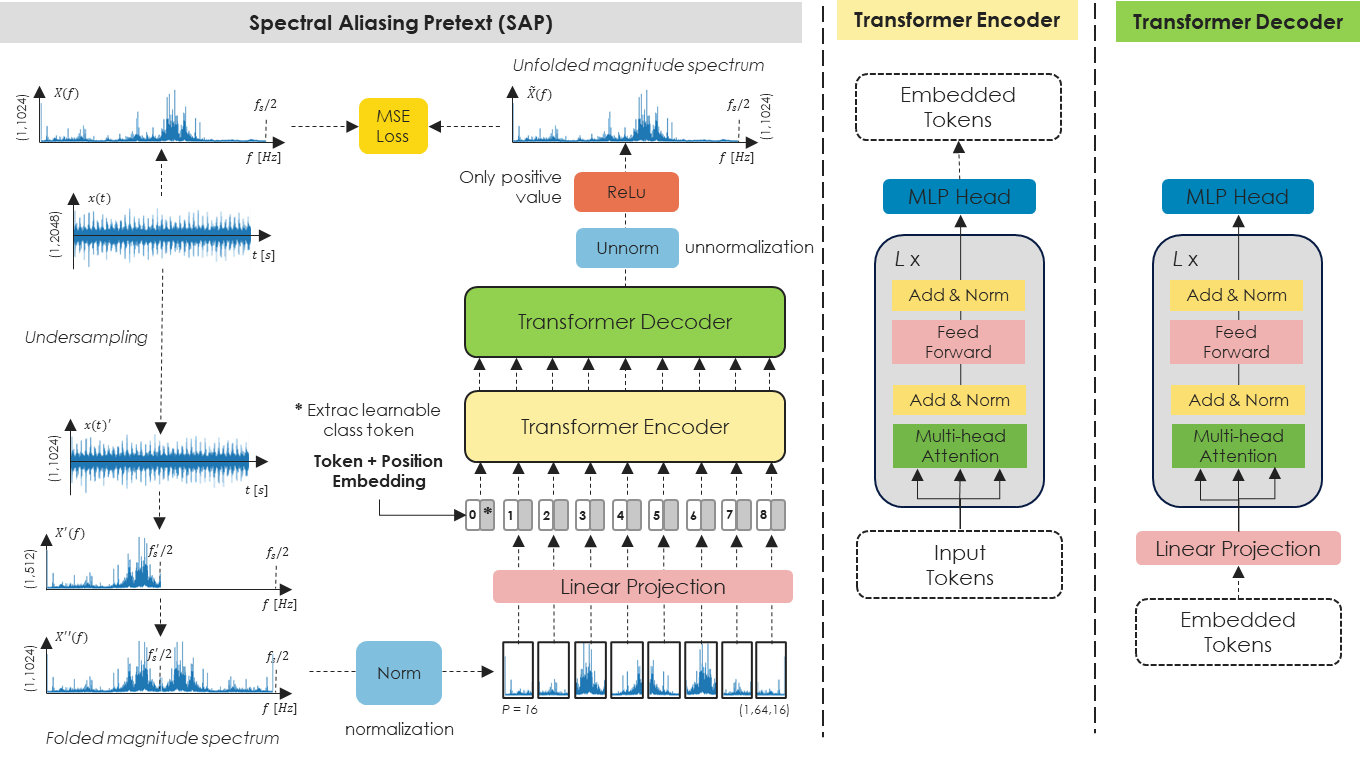}
    
    \caption{Spectral Aliasing Pretext (SAP) pretraining: from aliased spectrum generation to reconstruction. $f_s$ the original sample frequency, $f_s'$ the corrupted sample frequency }
    \label{fig:SAP}
\end{figure*}

Deep learning has transformed machinery fault diagnosis, achieving remarkable accuracy when large labeled datasets are available\cite{su_machine_2024}. However, in operational technology (OT) environments, collecting comprehensive labeled fault data is a major practical limitation\cite{li_small_2024}. Acquiring vibration signals under diverse fault conditions requires costly test benches, controlled degradation protocols, and expert identification\cite{noauthor_nasa_nodate} resources rarely available in industrial settings. This data scarcity creates a fundamental bottleneck for deploying deep learning in real-world predictive maintenance.

Self-supervised learning (SSL) offers a promising paradigm to address this challenge\cite{balestriero_cookbook_2023}. Instead of requiring expensive labels upfront, SSL methods first pretrain models on abundant unlabeled data using pretext tasks automatically learning without human supervision. For time series analysis, several SSL approaches have emerged in recent years. Contrastive methods like TS2Vec\cite{yue_ts2vec_2022} and TS-TCC\cite{eldele_time-series_2021} learn representations by maximizing agreement between augmented views of the same signal, drawing inspiration from successful computer vision frameworks like SimCLR\cite{chen_simple_2020}. Generative methods employ masked reconstruction strategies, predicting missing portions of the input similar to BERT's\cite{devlin_bert_2019} approach in natural language processing, or Masked Autoencoders (MAE)\cite{he_masked_2021} in computer vision. While Masked Image Modeling (MIM)\cite{hondru_masked_2025} works well for images due to strong spatial redundancy, its relevance for vibration time series used in fault classification is less clear. In frequency-domain signals, diagnostic information is often concentrated in specific harmonics and in the global relationships between frequency components. Random masking may remove or keep these components without control, allowing the model to reconstruct the signal using local correlations without necessarily learning the physical structure of fault signatures. Although these methods have shown promising results for general time-series representation learning and have started to be explored for machinery fault diagnosis \cite{rombach_contrastive_2021}. They do not explicitly exploit the deterministic spectral organization produced by mechanical faults.

Recent work on semi-supervised and transfer learning for bearing diagnosis\cite{eldele_unifault_2025} has made progress in reducing labeled data requirements, but these approaches still do not exploit frequency-domain physics as a core learning principle. However, current SSL methods face a critical limitation when applied to vibration-based fault diagnosis. Most approaches operate primarily in the time domain and rely heavily on data augmentation strategies adding noise, scaling amplitudes, temporal cropping, or applying random transformations\cite{zhou_faultformer_2024}. Furthermore, existing methods do not explicitly leverage the rich physical structure inherent to the frequency domain, where mechanical faults manifest as deterministic, mathematically predictable patterns. 

We argue that for vibration analysis, working directly with frequency-domain representations offers distinct advantages. Mechanical faults generate vibrations at characteristic frequencies determined by geometry and kinematics: a bearing defect produces periodic impacts whose frequencies depend on the bearing dimensions, number of rolling elements, and shaft speed. These deterministic signatures are most clearly visible in the magnitude spectrum. Moreover, the measured vibration signal exhibits a dual nature a superposition of cyclostationary components (periodic patterns with deterministic frequencies associated with faults) and stochastic modulation effects (structural resonances and propagation paths that vary randomly). The Fourier transform naturally separates these components, projecting the signal into a space where fault-related features are compact and discriminant\cite{antoni_cyclic_2007}.

Analyzing these frequency patterns requires capturing long-range dependencies across the spectrum. Fault signatures often appear as low-frequency fundamental peaks followed by high-frequency modulation sidebands distributed across wide frequency bands. Traditional convolutional neural networks (CNNs), which have shown success in supervised vibration-based fault diagnosis\cite{lu_intelligent_2017}, are designed to exploit local spatial patterns and struggle with these distributed relationships. The Transformer architecture, with its global self-attention mechanism, naturally models dependencies between frequency components. Recent work has demonstrated the effectiveness of Transformers for time series classification\cite{wen_transformers_2023} and has begun to explore their application to vibration-based condition monitoring though primarily in supervised settings with abundant labeled data.

We propose a SSL approach that addresses these limitations. Our core contribution is Spectral Aliasing Pretext (SAP), a novel pretext task that operates directly on magnitude spectra and exploits a fundamental phenomenon in signal processing: spectral aliasing. When a signal is undersampled, high-frequency components fold back into the low-frequency range following deterministic mathematical rules, creating an ambiguous but structured corruption. We train a Transformer encoder-decoder to reverse this folding to reconstruct the non-aliased spectrum from an intentionally aliased version. This task compels the model to learn deep physical constraints about spectral structure without relying on potentially destructive augmentations. Existing SSL methods do not exploit aliasing in the amplitude spectrum as a pretext task, which represents a gap in leveraging frequency-domain invariants for fault diagnosis. Our main contributions are threefold:

\begin{enumerate}
    \item We propose \textbf{Spectral Aliasing Pretext (SAP)}, a self-supervised pretraining method based on spectral aliasing, a signal processing phenomenon for vibration data.
    
    \item We show that a Transformer trained on magnitude spectra with SAP learns effective representations, reaching near-optimal fault classification performance on CWRU using only a small amount of labeled data.
    
\end{enumerate}

The remainder of this paper is organized as follows. Section II describes the datasets and our proposed methodology, including the mathematical formulation of spectral aliasing and the Transformer architecture. Section III details the experimental protocol for pretraining and downstream evaluation. Section IV presents comprehensive results on classification performance, followed by discussion of the learned representations. Section V concludes with perspectives on future research directions.

\section{Materiel and Methods}

\subsection{Datasets}

We use the widely adopted Case Western Reserve University (CWRU) bearing dataset as a evaluation benchmark. It contains vibration signals acquired under four bearing health conditions: \textit{normal}, \textit{inner-race fault}, \textit{outer-race fault}, and \textit{ball fault}. Fault diameters range from 0.007 to 0.028 inches, covering increasing levels of defect severity. Signals were collected at several load levels from 0 to 3 horsepower (motor speeds of 1797 to 1730 RPM). We use only the Fan End accelerometer sampled at 12 kHz. The dataset provides a controlled environment in which spectral structures, including characteristic fault harmonics are well established. 

\subsection{Proposed Approach}
\subsubsection{Problem Formulation}
The problem of spectral un-folding is fundamentally rooted in spectral aliasing, which occurs when a continuous time-domain signal, $x(t)$, is sampled at a frequency $f_s$ that is less than twice its maximum frequency component, thereby violating the Nyquist-Shannon sampling theorem. If the signal is undersampled, the new sampling frequency $f'_s$ is reduced. Original frequency components $f > f'_s/2$ are then folded back into the valid range $[0, f'_s/2]$. Mathematically, the aliased spectrum $X'(f)$ is represented as a superposition of the true spectrum $X(f)$ and its replicas shifted by multiples of the reduced sampling rate Eq. \eqref{eq:aliasing problem}.

\begin{equation} 
\label{eq:aliasing problem} 
X'(f) = \sum_{k=-\infty}^{\infty} X(f - kf'_s)
\end{equation}

When magnitude spectrum is estimated using the Fast Fourier Transform (FFT) on the undersampled signal, this summation manifests as folding, where high-frequency information is superimposed onto low-frequency information. The theoretical challenge is to reverse this aliasing that is, to infer and reconstruct the uncorrupted frequency distribution, from the mathematically ambiguous folded spectrum.

\subsubsection{Pretext Task}

We construct a self-supervised pretext task explicitly designed to capture the non-linear relationship between aliased and non-aliased spectral components. The complete preprocessing and corruption pipeline involves six sequential steps: windowing, undersampling, mean centering, discret Fourier transform, symmetric input preparation and normalization.

\textbf{Window Slicing.} Raw time-domain vibration signals are first segmented using a sliding window of $N = 2048$ points with a stride of 256 points. This produces overlapping temporal segments $x(t) \in \mathbb{R}^{2048}$ that capture local vibrational behavior.

\textbf{Undersampling.} To create the corrupted input for our pretext task, each raw segment $x(t)$ is deliberately undersampled by a factor of two. This produces a corrupted segment $x'(t) \in \mathbb{R}^{1024}$. By reducing the Nyquist frequency from $f_s/2$ to $f_s/4$, this undersampling causes spectral aliasing range according to the mathematical relationship described in Eq.~\eqref{eq:aliasing problem} .

\textbf{Mean Centering.} Both the original segment $x(t)$ and the corrupted segment $x'(t)$ are mean-centered to remove the continu component. This ensures that the spectrum focuses on oscillatory components rather than constant offsets.

\textbf{Discret Fourier Transform.} We apply the real-valued Fast Fourier Transform to both centered segments, transforming them from the time domain to the frequency domain. Since the input signals are real-valued, we exploit the Hermitian symmetry of the finite Discrete Fourier Transform (DFT) to compute only the non-redundant positive-frequency components. 

\textbf{Symmetric Input Preparation.} The undersampled spectrum $X'(f) \in \mathbb{R}^{N/4}$ has half the dimensionality of the target spectrum $X(f) \in \mathbb{R}^{N/2}$, creating a dimensional mismatch for the Transformer encoder-decoder architecture, which requires identical sequence lengths at input and output. To restore the required dimensional consistency, we construct an extended input $X''(f)$ by concatenating $X'(f)$ with its spatially reversed version. 

\textbf{Normalization.} To stabilize training, magnitude spectra are log-compressed and standardized using global statistics ($\mu_X, \sigma_X$) from the pretraining dataset:

\begin{equation}
\label{eq:normalization_new}
X_i(f)_{\text{norm}} = \frac{log(1+X_i(f)) - \mu_X}{\sigma_X} , \quad i \in \{0, \ldots, n-1\}
\end{equation}

Unlike per-sample scaling, using global statistics ensures consistent input distributions and improves robustness to amplitude variations across operating conditions. Preprocessed dataset statistics are summarized in Table~\ref{tab:dataset_stats}. 

\begin{table}
    \centering
    \caption{Dataset statistics after preprocessing. No Sample refers to the number of  normalized magnitude spectrum.}
    \begin{tabular}{cccc}\toprule
         Dataset&   Sample Frequency $f_s$& No. Sample &No. Classes\\\midrule
         CWRU&   12 kHz& 27593 &10\\ \bottomrule
         % LASPI&   25,6 kHz& 285696 &6\\ \bottomrule
    \end{tabular}
    \label{tab:dataset_stats}
\end{table}

\begin{figure*}[t]
    \centering
    \includegraphics[width=1\linewidth]{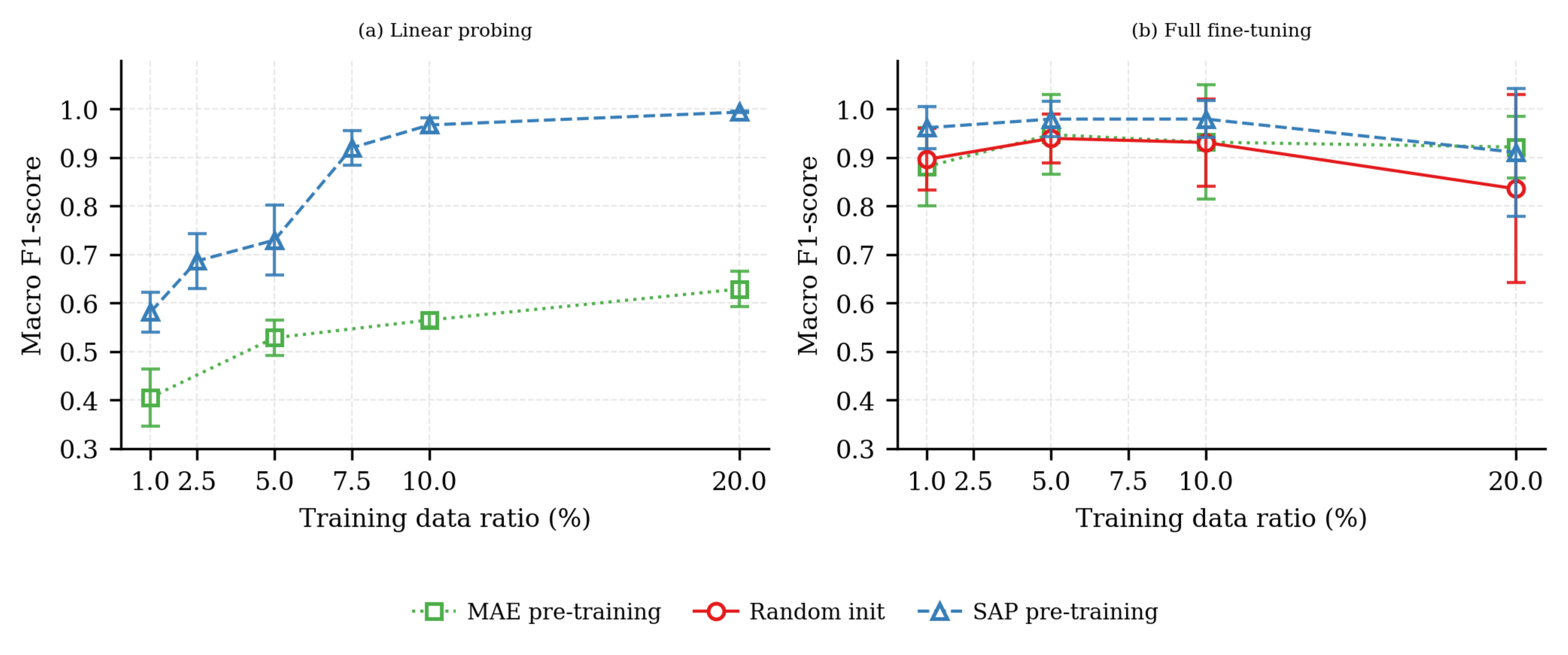}
    
    \caption{Macro F1-score as a function of the training data ratio for (a) linear probing and (b) full fine-tuning on CWRU. Results are reported for random initialization, SAP pre-training, and MAE pre-training. The masking ratio for MAE is 40\%. }
    \label{fig:CWRU_results}
\end{figure*}

\subsubsection{Model architecture}

The model architecture (Fig.~\ref{fig:SAP}) is based on a symmetric Transformer encoder-decoder structure. The input spectrum is tokenized into patches of size $P$. The token is projected in embedding dimension fixed at $d_{model}$. Both the encoder and decoder consist of $L$ identical layers, and each layer employs multi-head self-attention with $H$ heads. A learnable class token and Sinusoidal Positional Embeddings are used to maintain sequence integrity. The entire structure is regularized using a dropout rate of $\rho_{drop}$. For the downstream classification task, the class token from the encoder (the backbone) is passed to a linear  head classifier.
\begin{table}[h]
\centering
\caption{Transformer encoder-decoder hyperparameters}
\label{tab:hyperparameters}
\begin{tabular}{lcc}
\toprule
\textbf{Parameter} & \textbf{Variable} & \textbf{Value} \\
\midrule
Patch Size & $P$ & 16 \\
Hidden Dimension & $d_{model}$ & 512 \\
Number of Attention Heads & $H$ & 8 \\
Number of Layers (E/D) & $L$ & 3 \\
Dropout Rate & $\rho_{drop}$ & $2.57 \times 10^{-1}$\\
Learning Rate & $\alpha$ & $3.67 \times 10^{-4}$\\
Weight Decay & $\lambda$ & $1.11 \times 10^{-5}$\\
Batch Size & $B$ & 64 \\
\bottomrule
\end{tabular}
\end{table}

\section{Experiments}

\subsection{Dataset Split Strategy}
Two distinct data splitting strategies are employed depending on the learning stage. For the self-supervised pre-training, the CWRU dataset is split using a heterogeneous strategy, where all available spectra are randomly partitioned into training and validation sets. A seed is used to ensure reproducibility across all experiments. We acknowledge that the random partition used during self-supervised pre-training may introduce overlap between training and validation spectra due to the sliding-window procedure. However, the downstream test sets are built from operating conditions strictly excluded from training (unseen speed or speed-load combinations), ensuring an independent evaluation protocol and preventing leakage between training and test domains. For the downstream supervised tasks, the data are split according to the operating speeds. For all experiments, the backbone is pre-trained on CWRU and specialized on CWRU using training data collected at 1750, 1772, and 1797 rpm, while evaluation is performed at 1730 rpm. 

\subsection{Pre-training Setup}
Self-supervised pre-training is performed exclusively on the CWRU dataset. Two pre-training strategies are considered: the proposed SAP approach and a Masked Autoencoder (MAE) \cite{he_masked_2021} used as a baseline for comparison. The model is trained to minimize the mean squared reconstruction error (MSE). This choice reflects industrial scenarios where operating conditions may be unknown, and highlights the ability of self-supervised learning to learn robust condition representations.

\subsection{Downstream Setup}
The downstream task consists of supervised fault classification and is evaluated on both the CWRU dataset. The Transformer encoder learned during the pre-training stage is used as a feature extraction backbone. To assess label efficiency, we use different ratio of the downstream training set is used for supervised learning. Two training regimes are considered for evaluate the models :

\textbf{Linear probing}: the backbone is initialized with self-supervised pre-trained weights and kept frozen, while a linear classification head is trained on top of the extracted features.

\textbf{Full fine-tuning}: the backbone is initialized with self-supervised pre-trained weights and jointly optimized with the linear classification head.

We compare against a supervised baseline where the backbone is randomly initialized and trained on the available labeled data. Performance is evaluated using the F1-score on the held-out operating condition, ensuring a fair comparison between self-supervised and supervised methods under limited data and domain shifts.

\section{Results and Discussion}

\begin{table}[!t]
  \caption{Effect of Mask Ratio on Linear Probing and Full Fine-tuning Performance (\%)}
  \label{tab:mask_ratio}
  \centering
  \begin{tabular}{ccc}
    \hline
    \textbf{Mask Ratio} & \textbf{Linear Probing} & \textbf{Full Fine-tune} \\
    \hline
    10\% & 58.6 & 89.5 \\
    20\% & 54.1 & 81.5 \\
    30\% & 41.5 & 82.4 \\
    40\% & \textbf{62.9} & \textbf{92.2} \\
    50\% & 53.2 & 76.3 \\
    60\% & 33.6 & 88.3 \\
    70\% & 50.8 & 82.1 \\
    75\% & 42.2 & 90.1 \\
    80\% & 48.8 & 81.5 \\
    90\% & 48.1 & 88.5 \\
    \hline
  \end{tabular}
\end{table}

We evaluate the proposed Spectral Aliasing Pretext (SAP) on downstream fault classification under varying proportions of labeled data. Performance is reported using the macro F1-score on a held-out operating condition.

In the linear probing setting (Fig. \ref{fig:CWRU_results}), SAP consistently outperforms MAE pretraining across all training data ratios. SAP rapidly converges and reaches a macro F1-score close to 1.0 with only $20\%$ of the training set while maintaining very low variance across runs. Since the backbone is frozen in this setting, these results directly reflect the intrinsic quality of the learned representations. The strong performance combined with low variance suggests that SAP learns features that are both highly discriminative and well structured in feature space. In other words, samples from different fault classes become linearly separable with minimal supervision.

This behavior contrasts with MAE pretraining, which shows lower performance and higher variability. The effectiveness of the MAE baseline is also notably sensitive to its masking hyperparameter (Table~\ref{tab:mask_ratio}.), the macro F1-score for both linear probing and full fine-tuning reaches its peak at a mask ratio of $40\%$. Masking-based reconstruction mainly encourages the model to learn local correlations between frequency bins, but does not explicitly enforce the learning of global spectral relationships. As a result, the learned features remain less structured and require additional supervised adaptation to become discriminative. The stability of SAP in linear probing indicates that the pretext task itself provides a strong inductive bias: reconstructing an aliased spectrum requires modeling global relationships and dependencies across frequencies. These constraints appear to guide the model toward a representation space that is already aligned with fault classes before any supervised training.

In the full fine-tuning setting (Fig. \ref{fig:CWRU_results}), all methods, random initialization (full supervised), MAE pretraining, and SAP pretraining reach similar average performance, without significant differences. However, the optimization process is less stable, with higher variance across runs. Surprisingly, even fully supervised training from random initialization does not consistently converge toward stable results, and fine-tuning tends to increase performance variability. This may be explained by the large number of parameters updated during fine-tuning, which makes optimization more sensitive to initialization and sampling effects. It may also reflect the relative simplicity of the CWRU dataset. Many models can eventually fit the data, but the optimization trajectory remains unstable. On more complex industrial data, full fine-tuning may become beneficial, but the strong stability observed in linear probing indicates that SAP learns robust representations.

Overall, the most informative comparison is therefore obtained in the linear probing regime, where representation quality can be assessed independently of full optimization. In this setting, SAP produces stable and linearly separable features that enable near-perfect classification with limited labeled data. The combination of high accuracy, rapid convergence, and low variance suggests that pretext tasks such as spectral aliasing can lead to more robust and better-structured representations for vibration-based fault diagnosis.

\section{Conclusion and Perspective}

Experimental results on the CWRU dataset show that SAP
produces more discriminative representations than masking-
based pretraining, achieving strong classification performance
with only a small fraction of labeled data. The method provides
a stable initialization for downstream learning and improves
label efficiency compared with both random initialization
and MAE pretraining. These results highlight the benefit of
integrating signal processing knowledge into self-supervised
learning for industrial condition monitoring. Future work will
extend the comparison to additional SSL methods such as
TS2Vec and TS-TCC, investigate cross-dataset generalization,
analyze the impact of potential data leakage induced by
overlapping window segmentation, and further justify key
design choices including the symmetric spectrum concatenation
strategy.

\section*{Acknowledgment}
This work was granted access to the HPC resources of IDRIS under the allocation 20XX-AD010114820R2 made by GENCI.

\scriptsize

\bibliographystyle{ieeetr} 
\bibliography{ref} 

@article{yue_ts2vec_2022,
	title = {{TS}2Vec: Towards Universal Representation of Time Series},
	volume = {36},
	rights = {Copyright (c) 2022 Association for the Advancement of Artificial Intelligence},
	issn = {2374-3468},
	url = {https://ojs.aaai.org/index.php/AAAI/article/view/20881},
	doi = {10.1609/aaai.v36i8.20881},
	shorttitle = {{TS}2Vec},
	pages = {8980--8987},
	number = {8},
	journaltitle = {Proceedings of the {AAAI} Conference on Artificial Intelligence},
	author = {Yue, Zhihan and Wang, Yujing and Duan, Juanyong and Yang, Tianmeng and Huang, Congrui and Tong, Yunhai and Xu, Bixiong},
	urldate = {2024-11-18},
	date = {2022-06-28},
	langid = {english},
	note = {Number: 8},
}

@misc{balestriero_cookbook_2023,
	title = {A Cookbook of Self-Supervised Learning},
	url = {http://arxiv.org/abs/2304.12210},
	doi = {10.48550/arXiv.2304.12210},
	number = {{arXiv}:2304.12210},
	publisher = {{arXiv}},
	author = {Balestriero, Randall and Ibrahim, Mark and Sobal, Vlad and Morcos, Ari and Shekhar, Shashank and Goldstein, Tom and Bordes, Florian and Bardes, Adrien and Mialon, Gregoire and Tian, Yuandong and Schwarzschild, Avi and Wilson, Andrew Gordon and Geiping, Jonas and Garrido, Quentin and Fernandez, Pierre and Bar, Amir and Pirsiavash, Hamed and {LeCun}, Yann and Goldblum, Micah},
	urldate = {2025-02-18},
	date = {2023-06-28},
	eprinttype = {arxiv},
	eprint = {2304.12210 [cs]},
}

@misc{chen_simple_2020,
	title = {A Simple Framework for Contrastive Learning of Visual Representations},
	url = {http://arxiv.org/abs/2002.05709},
	doi = {10.48550/arXiv.2002.05709},
	number = {{arXiv}:2002.05709},
	publisher = {{arXiv}},
	author = {Chen, Ting and Kornblith, Simon and Norouzi, Mohammad and Hinton, Geoffrey},
	urldate = {2025-02-19},
	date = {2020-07-01},
	langid = {english},
	eprinttype = {arxiv},
	eprint = {2002.05709 [cs]},
}

@article{zhou_faultformer_2024,
	title = {{FaultFormer}: Pretraining Transformers for Adaptable Bearing Fault Classification},
	volume = {12},
	issn = {2169-3536},
	url = {http://arxiv.org/abs/2312.02380},
	doi = {10.1109/ACCESS.2024.3399670},
	shorttitle = {{FaultFormer}},
	pages = {70719--70728},
	journaltitle = {{IEEE} Access},
	shortjournal = {{IEEE} Access},
	author = {Zhou, Anthony and Farimani, Amir Barati},
	urldate = {2025-06-27},
	date = {2024},
	eprinttype = {arxiv},
	eprint = {2312.02380 [cs]},
}

@misc{eldele_unifault_2025,
	title = {{UniFault}: A Fault Diagnosis Foundation Model from Bearing Data},
	url = {http://arxiv.org/abs/2504.01373},
	doi = {10.48550/arXiv.2504.01373},
	shorttitle = {{UniFault}},
	number = {{arXiv}:2504.01373},
	publisher = {{arXiv}},
	author = {Eldele, Emadeldeen and Ragab, Mohamed and Qing, Xu and Edward and Chen, Zhenghua and Wu, Min and Li, Xiaoli and Lee, Jay},
	urldate = {2025-10-06},
	date = {2025-04-02},
	langid = {english},
	eprinttype = {arxiv},
	eprint = {2504.01373 [cs]},
}

@misc{eldele_time-series_2021,
	title = {Time-Series Representation Learning via Temporal and Contextual Contrasting},
	url = {http://arxiv.org/abs/2106.14112},
	doi = {10.48550/arXiv.2106.14112},
	number = {{arXiv}:2106.14112},
	publisher = {{arXiv}},
	author = {Eldele, Emadeldeen and Ragab, Mohamed and Chen, Zhenghua and Wu, Min and Kwoh, Chee Keong and Li, Xiaoli and Guan, Cuntai},
	urldate = {2026-01-05},
	date = {2021-06-26},
	eprinttype = {arxiv},
	eprint = {2106.14112 [cs]},
}

@misc{devlin_bert_2019,
	title = {{BERT}: Pre-training of Deep Bidirectional Transformers for Language Understanding},
	url = {http://arxiv.org/abs/1810.04805},
	doi = {10.48550/arXiv.1810.04805},
	shorttitle = {{BERT}},
	number = {{arXiv}:1810.04805},
	publisher = {{arXiv}},
	author = {Devlin, Jacob and Chang, Ming-Wei and Lee, Kenton and Toutanova, Kristina},
	urldate = {2026-01-05},
	date = {2019-05-24},
	eprinttype = {arxiv},
	eprint = {1810.04805 [cs]},
}

@article{lu_intelligent_2017,
	title = {Intelligent fault diagnosis of rolling bearing using hierarchical convolutional network based health state classification},
	volume = {32},
	issn = {1474-0346},
	url = {https://www.sciencedirect.com/science/article/pii/S1474034616301148},
	doi = {10.1016/j.aei.2017.02.005},
	pages = {139--151},
	journaltitle = {Advanced Engineering Informatics},
	shortjournal = {Advanced Engineering Informatics},
	author = {Lu, Chen and Wang, Zhenya and Zhou, Bo},
	urldate = {2026-01-05},
	date = {2017-04-01},
}

@article{antoni_cyclic_2007,
	title = {Cyclic spectral analysis of rolling-element bearing signals: Facts and fictions},
	volume = {304},
	issn = {0022-460X},
	url = {https://www.sciencedirect.com/science/article/pii/S0022460X07001551},
	doi = {10.1016/j.jsv.2007.02.029},
	shorttitle = {Cyclic spectral analysis of rolling-element bearing signals},
	pages = {497--529},
	number = {3},
	journaltitle = {Journal of Sound and Vibration},
	shortjournal = {Journal of Sound and Vibration},
	author = {Antoni, J.},
	urldate = {2026-01-05},
	date = {2007-07-24},
}

@article{su_machine_2024,
	title = {Machine Learning Approaches for Diagnostics and Prognostics of Industrial Systems Using Open Source Data from {PHM} Data Challenges: A Review},
	volume = {15},
	rights = {Copyright (c) 2024 International Journal of Prognostics and Health Management},
	issn = {2153-2648},
	url = {https://papers.phmsociety.org/index.php/ijphm/article/view/3993},
	doi = {10.36001/ijphm.2024.v15i2.3993},
	shorttitle = {Machine Learning Approaches for Diagnostics and Prognostics of Industrial Systems Using Open Source Data from {PHM} Data Challenges},
	number = {2},
	journaltitle = {International Journal of Prognostics and Health Management},
	author = {Su, Hanqi and Lee, Jay},
	urldate = {2026-01-27},
	date = {2024-09-17},
	langid = {english},
}

@article{li_small_2024,
	title = {Small data challenges for intelligent prognostics and health management: a review},
	volume = {57},
	issn = {1573-7462},
	url = {https://doi.org/10.1007/s10462-024-10820-4},
	doi = {10.1007/s10462-024-10820-4},
	shorttitle = {Small data challenges for intelligent prognostics and health management},
	pages = {214},
	number = {8},
	journaltitle = {Artificial Intelligence Review},
	shortjournal = {Artif Intell Rev},
	author = {Li, Chuanjiang and Li, Shaobo and Feng, Yixiong and Gryllias, Konstantinos and Gu, Fengshou and Pecht, Michael},
	urldate = {2026-01-27},
	date = {2024-07-23},
	langid = {english},
}

@online{noauthor_nasa_nodate,
	title = {{NASA} Prognostics Center of Excellence Data Set Repository [Mirror]},
	url = {https://data.phmsociety.org/nasa/},
	titleaddon = {{PHM} Society Data Repository},
	urldate = {2026-01-27},
	langid = {american},
}

@article{rombach_contrastive_2021,
	title = {Contrastive Learning for Fault Detection and Diagnostics in the Context of Changing Operating Conditions and Novel Fault Types},
	volume = {21},
	rights = {http://creativecommons.org/licenses/by/3.0/},
	issn = {1424-8220},
	url = {https://www.mdpi.com/1424-8220/21/10/3550},
	doi = {10.3390/s21103550},
	pages = {3550},
	number = {10},
	journaltitle = {Sensors},
	publisher = {Multidisciplinary Digital Publishing Institute},
	author = {Rombach, Katharina and Michau, Gabriel and Fink, Olga},
	urldate = {2026-01-27},
	date = {2021-01},
	langid = {english},
}

@misc{wen_transformers_2023,
	title = {Transformers in Time Series: A Survey},
	url = {http://arxiv.org/abs/2202.07125},
	doi = {10.48550/arXiv.2202.07125},
	shorttitle = {Transformers in Time Series},
	number = {{arXiv}:2202.07125},
	publisher = {{arXiv}},
	author = {Wen, Qingsong and Zhou, Tian and Zhang, Chaoli and Chen, Weiqi and Ma, Ziqing and Yan, Junchi and Sun, Liang},
	urldate = {2026-01-27},
	date = {2023-05-11},
	eprinttype = {arxiv},
	eprint = {2202.07125 [cs]},
}

@misc{he_masked_2021,
	title = {Masked Autoencoders Are Scalable Vision Learners},
	url = {http://arxiv.org/abs/2111.06377},
	doi = {10.48550/arXiv.2111.06377},
	number = {{arXiv}:2111.06377},
	publisher = {{arXiv}},
	author = {He, Kaiming and Chen, Xinlei and Xie, Saining and Li, Yanghao and Dollár, Piotr and Girshick, Ross},
	urldate = {2025-06-27},
	date = {2021-12-19},
	eprinttype = {arxiv},
	eprint = {2111.06377 [cs]},
}

@misc{hondru_masked_2025,
	title = {Masked Image Modeling: A Survey},
	url = {http://arxiv.org/abs/2408.06687},
	doi = {10.48550/arXiv.2408.06687},
	shorttitle = {Masked Image Modeling},
	number = {{arXiv}:2408.06687},
	publisher = {{arXiv}},
	author = {Hondru, Vlad and Croitoru, Florinel Alin and Minaee, Shervin and Ionescu, Radu Tudor and Sebe, Nicu},
	urldate = {2026-02-11},
	date = {2025-07-10},
	eprinttype = {arxiv},
	eprint = {2408.06687 [cs]},
}

\end{document}